\documentclass[10pt,twocolumn,letterpaper]{article}

\usepackage{cvpr}              

\usepackage[dvipsnames]{xcolor}

\usepackage[percent]{overpic}
\usepackage{comment}
\usepackage{arydshln}

\definecolor{cvprblue}{rgb}{0.21,0.49,0.74}
\usepackage[pagebackref,breaklinks,colorlinks,citecolor=cvprblue]{hyperref}

\def\paperID{248} 
\def\confName{3DV\xspace}
\def\confYear{2026\xspace}

\title{Diverse Signer Avatars with Manual and Non-Manual Feature Modelling for Sign Language Production}

\author{Mohamed Ilyes Lakhal, Richard Bowden\\
CVSSP, University of Surrey, Guildford, United Kingdom\\
{\tt\small \{m.lakhal, r.bowden\}@surrey.ac.uk}}

\begin{document}
\twocolumn[{%
\renewcommand\twocolumn[1][]{#1}%
\maketitle
\includegraphics[trim={0 0px 0 0},clip,width=.9\linewidth]{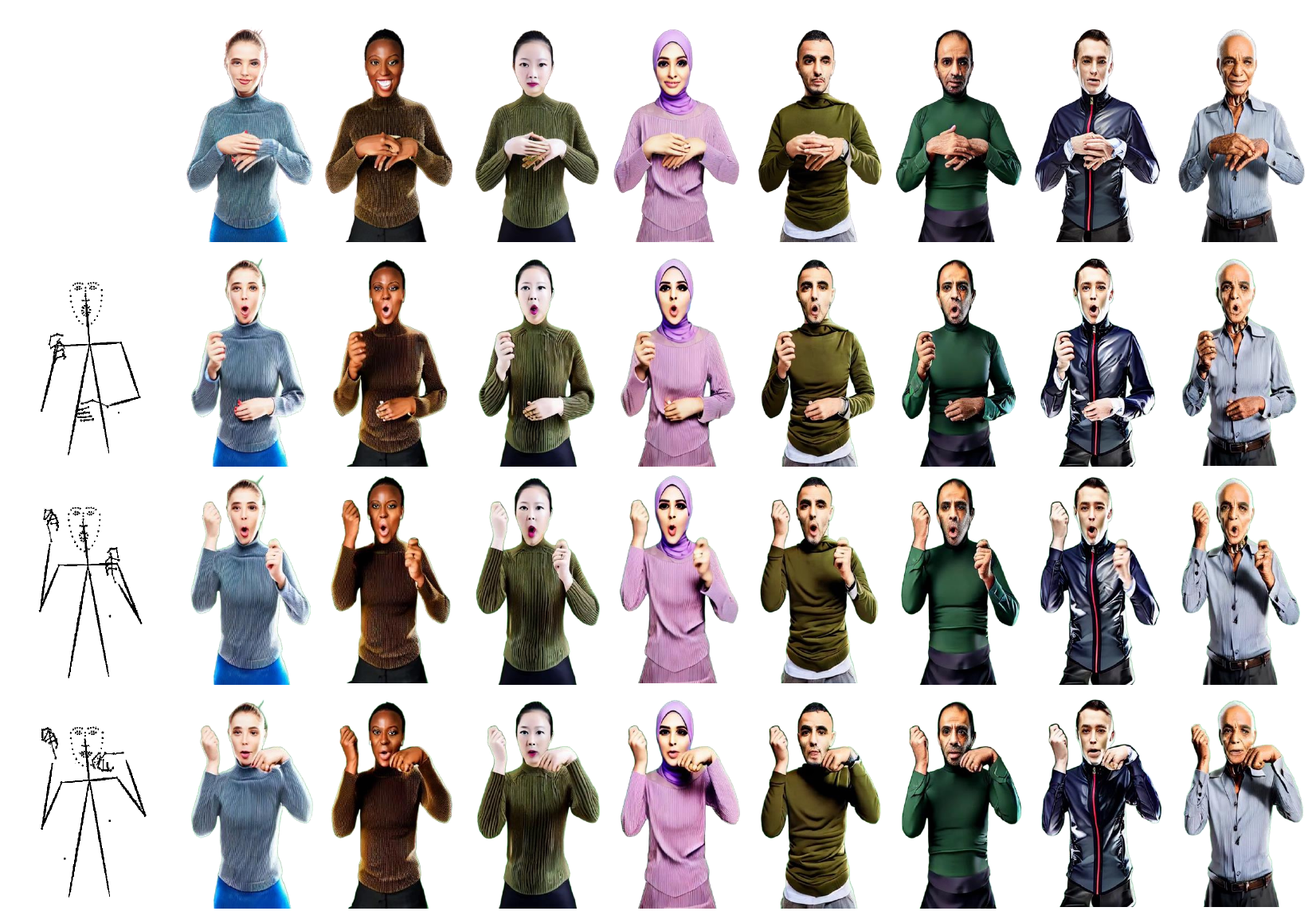}
\vspace{-.5em}
\captionof{figure}{Our goal is to generate photorealistic digital avatars that preserve essential sign language cues — such as facial expressions, hand movements, and mouthing — while enabling diversity across ethnicities and adaptability to different sign languages. %
}
\vspace{10pt}
\label{fig:teaser}

}]

\begin{abstract}
The diversity of sign representation is essential for Sign Language Production (SLP) as it captures variations in appearance, facial expressions, and hand movements. However, existing SLP models are often unable to capture diversity while preserving visual quality and modelling non-manual attributes such as emotions. To address this problem, we propose a novel approach that leverages Latent Diffusion Model (LDM) to synthesise photorealistic digital avatars from a generated reference image. We propose a novel sign feature aggregation module that explicitly models the non-manual features (\textit{e.g.}, the face) and the manual features (\textit{e.g.}, the hands). We show that our proposed module ensures the preservation of linguistic content while seamlessly using reference images with different ethnic backgrounds to ensure diversity. Experiments on the YouTube-SL-25 sign language dataset show that our pipeline achieves superior visual quality compared to state-of-the-art methods, with significant improvements on perceptual metrics.
\end{abstract}    
\section{Introduction}
\label{sec:intro}

One of the advantages of written text is that it enables broad accessibility without being dependent on the personal characteristics of the writer. When someone writes a blog, the message is conveyed regardless of the identity of the author. However, this is not usually the case with a language such as sign language. The visual nature of sign language inherently reflects the appearance and style of the signer. This manuscript addresses the timely and relevant problem of how to represent sign language in a way that captures the important subtleties of communication while allowing for diverse signer representations.
At a time when accessibility is on the rise, it is important that more content is translated into national sign languages. With the increasing national recognition of sign languages in countries around the world, the expansion of accessible content is crucial. But whose representation should appear in this content?

There is a growing interest in generating sign language materials that feature a diverse range of signer representations, ensuring inclusivity while preserving the linguistic richness and integrity of the message. To address this problem, we propose to explicitly model the most important cues in each sign sequence, namely: manual and non-manual features~\cite{Sandler_Lillo-Martin_2006}. In the context of sign language, manual features are represented by hand and arm movements that convey the primary lexical content of the signs, while non-manual features include facial expressions, head movements and body postures that provide grammatical and emotional context~\footnote{Although non-manual features convey important lexical and grammatical information through facial expressions, as well as subtle movements of the arms, shoulders, and body, it is generally recognised that most simplifications of non-manual modelling focus mainly on the facial region~\cite{6016973}.}.

Early deep learning approaches for sign language synthesis relied on Generative Adversarial Networks (GANs)~\cite{goodfellow2014generative}. While GANs proved effective in person-specific synthesis, they had difficulty generalising to unknown individuals from out-of- distribution samples as they often suffered from mode collapse, where the model produces only limited variations of outputs~\cite{pmlr-v70-arjovsky17a}. In contrast, recent advances in diffusion models have shown promise as they can be generalised to diverse reference images and allow for more flexible synthesis~\cite{9878449}. However, the direct application of these models to sign language synthesis often fails to adequately capture the manual (\textit{e.g.}, hand gestures) and non-manual (\textit{e.g.}, facial expressions) features that are essential for conveying meaning.

In this paper, we introduce \emph{Signer Avatar}, a framework that synthesises photorealistic signer avatars with rich demographic and ethnic diversity. Our work addresses a key concern of the deaf community: state-of-the-art models inadequately capture manual and non-manual features that are critical to sign language.
To explicitly model the manual and non-manual cues from a reference image, we develop a novel feature aggregation module based on 3D dilated convolutions. By using small dilation rates to capture fine facial details and larger rates to encode coarse hand and limb movements. Furthermore, we use Sapiens features~\cite{10.1007/978-3-031-73235-5_12} to regularise the aggregation between manual and non-manual cues, improving synthesis quality. 

In summary, our contributions are the following:

\begin{itemize}
    \item To our knowledge, this paper is the first to fully address the problem of generating photorealistic digital signer avatars with explicit emphasis on diversity. Our solution faithfully preserves the manual and non-manual features of the input video while enabling the animation of reference images from individuals with varied ethnic and demographic backgrounds.
    \item We present a novel multi‐scale spatio‐temporal feature aggregation module that combines pose, manual and non-manual feature modalities to capture the dynamics of sign language without encoding identity‐specific details.
    \item We propose a ControlNet‐based denoising network that combines foundation model visual embeddings along with the rich aggregated sign features, which together enable realistic, linguistically accurate and visually diverse avatar synthesis.
\end{itemize}
\section{Related Work}

\subsection{Sign language production} The ability to understand and produce sign language has long been a challenge for computer vision~\cite{Tamura_88}. Sign language production (SLP) is not about translation into a spoken or written language, but about synthesising coherent sign sequences from an input (e.g., text or semantics) that result in a fluent, natural sign output. SLP systems need to capture how manual and non-manual features are combined to convey meaning and then reproduce that meaning in sign form. Two main production paradigms have emerged in the literature: Text-to-Pose~\cite{9093516,saunders2021continuous}, which predicts skeletal motion sequences from text, and digital sign avatars~\cite{KARPOUZIS200754,hu2024expressive}, which produce fully articulated, photorealistic sign videos.

\textbf{Text‐to‐Pose.} Text‐to‐pose approaches learn a direct mapping from written or glossed source text to a sequence of skeletal joint positions (2D or 3D) representing the target sign~\cite{hwang2021non}. Early methods treated each word or gloss as a discrete token and generated a fixed‐length pose snippet via lookup tables or simple neural decoders~\cite{9093516}. More recent work abandons fixed frames and instead uses sequence‐to‐sequence architectures — transformer variants — with an explicit end‐of‐sequence to produce a variable-length output~\cite{saunders2020progressive,saunders2021continuous}.  Pose-based models generalise well across signer identity and camera viewpoints, but they often struggle to capture non‑manual markers (\textit{e.g.}, facial expressions and head movements) that convey important grammatical information.

\textbf{Digital Sign Avatars.} Digital Sign Avatar Pipelines aim to create animated digital human avatars signing. Previous methods that have proposed using 3D avatars to represent human signing usually rely on predefined rules, \textit{i.e.} pre‐animated sign segments are blended according to a phrase-level grammar and stored in a phrase lexicon~\cite{KARPOUZIS200754}. Although these can generate visually plausible signs, they require extensive manual labelling and cannot be easily adapted to novel sentences. First attempts with variational inference~\cite{Blei_2017} and generative adversarial networks~\cite{goodfellow2014generative} propose to force the distribution of the sign sample to follow a Gaussian distribution~\cite{10581951, Saunders_2021_FG,Silveira_2022_SIBGRAPI}. Although such an approach is promising for few individuals, these approaches fail to generate diverse individuals as they suffer from mode collapse~\cite{pmlr-v70-arjovsky17a} where the model ends up learning/collapsing to only a few patterns/modes. More recent works explored neural rendering techniques to increase visual realism in SLP~\cite{zhang2025guava,hu2024expressive,Lakhal_FG_25}. In particular, 3D Gaussian Splatting (3DGS), originally developed for static scene reconstruction, can be adapted for SLP through Skinned Multi-Person Linear (SMPL) which provides a canonical pose space and linear blending weights that allow 3DGS to animate a signer~\cite{Loper_2015_TOG}. Although this pipeline enables high-quality real-time synthesis, it remains limited by its dependence on person-specific geometry. As a result, 3DGS-based methods are excellent when trained for a single person, but they  struggle to generalise across multiple sign language users.

\subsection{Pose-guided human image synthesis} Pose-guided image synthesis attempts to generate a new view of a person by re-posing a single source image. Early approaches used a U-Net encoder decoder conditioned on 2D joint heatmaps~\cite{Ma_2017_NIPS} but struggled with part ambiguity and texture misalignment. To address this issue, later methods incorporated per-pixel semantic maps to distinguish body regions and control the transfer of high frequency detail~\cite{Dong_2018_NIPS}. DensePose UV- coordinates were also used to map image pixels onto a 3D surface, enabling more accurate placement of parts~\cite{Guler_2018_CVPR,Neverova_2018_ECCV}. The introduction of parametric body models such as SMPL enriches the prior with volumetric consistency and allows diverse poses in novel views~\cite{Li_2019_CVPR}.

Thanks to the rapid development of diffusion-based models~\cite{Ho_neurips_20,9878449}, significant progress has been made on this problem in generating more realistic and accurate human images that are conditioned on target poses. Denoising diffusion probabilistic models (DDPM)~\cite{Ho_neurips_20} was used to take a reference image and a target pose to gradually denoise the image of the person in the target pose in pixel space~\cite{bhunia2022pidm}. However, working directly with the pixel space as in DDPM is not suitable for high-resolution images. Therefore, the methods in~\cite{10377856, 10656324, shen2024advancingposeguidedimagesynthesis, shen2024imagpose, 10656410,zhang2025mimicmotion,peng2024controlnext,wang2025unianimate} use a latent diffusion model approach~\cite{9878449}, in which the reference image and the target pose are first encoded by a pre-trained (frozen) VAE encoder and the denoising is performed in feature space instead.

\begin{figure*}[t!]
    \centering
    \begin{overpic}[width=1.0\textwidth]{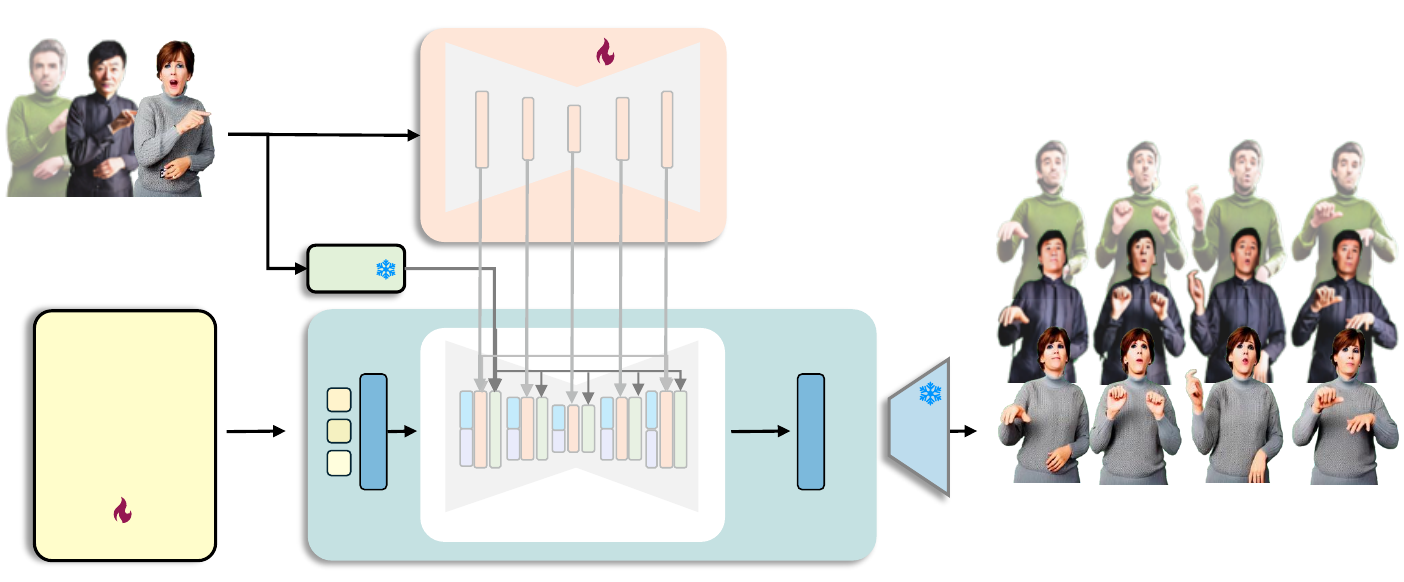}
    
    \put(38.0,37){ $\displaystyle \mathcal{E}_{\texttt{app}}$}

    \put(22.0,22.0){ $\displaystyle \mathcal{E}_{\texttt{feat}}$}

    \put(34.4,4.1){ Denoising U-Net}

    \put(54.0,4.1){ $\displaystyle \mu_{\theta}(z_t, t)$}

    \put(64.2,10.5){ $\displaystyle \mathcal{D}$}

    \put(3.3,14){  Sign features}
    \put(3.6,12.0){ aggregation}
    \put(5.1,9.9){  module}
    \put(4.6,7.8){ (Sec.~\ref{sec:aggregation})}

  \end{overpic}
    \caption{\textbf{Signer Avatar.} Our model explicitly processes the manual and non-manual features through our proposed sign feature aggregation module (Sec.~\ref{sec:aggregation}), which, together with the human keypoints and the abstract feature obtained from visual foundation model, forms the condition to our denoising network. The reference image is then combined through a denoising process to generate sign sequence.}
    \label{fig:main}
\end{figure*}

\section{Method}
We build on the latent diffusion model to generate photorealistic signer avatars capturing diverse demographics and ethnicities.
In Section~\ref{sub:method}, we present our proposed pipeline and explain how it adapts latent diffusion for diverse avatar synthesis. An overall representation of our pipeline can be seen in Figure~\ref{fig:main}.

\subsection{Overview}~\label{sub:method}
Our goal is to generate photorealistic and diverse signer videos using a Latent Diffusion Model (LDM)~\cite{9878449}. Given a sequence of RGB frames $\mathbf{I}_k$ defined on the lattice $\Omega = \{ 1, \dots, H \} \times \{ 1, \dots, W\}$, we aim to produce a photorealistic video sequence $\mathbf{I}_k^{\texttt{a}}$ of the corresponding sign from an unseen person.

Sign language sequences comprise manual features $\mathcal{M}$ (e.g., hand and torso keypoints) and non-manual features $\mathcal{N}$ (e.g., facial expressions). To ensure diversity, we leverage a Text-to-Image (T2I) model, such as ControlNet~\cite{zhang2023adding}, to produce a synthetic reference image $\mathbf{I}^{\texttt{r}}$. Our pipeline integrates $\mathcal{M}$, $\mathcal{N}$, and $\mathbf{I}^{\texttt{r}}$ to synthesise a frame $\mathbf{I}^{\texttt{a}}$, effectively mapping $\mathcal{M} \times \mathcal{N} \times \mathbf{I}^{\texttt{r}} \to \mathbf{I}^{\texttt{a}}$.

We build upon Stable Diffusion (SD)~\cite{9878449}, which extends Latent Diffusion Model (LDM), that operates on latent space instead of whole images to reduce computational burden. 
Given an image $\mathbf{x}$, we encode via an encoder $\mathcal{E}$ it into a latent representation such that $\mathbf{z} = \mathcal{E}(\mathbf{x})$. Likewise, the image is reconstructed through a decoder $\mathcal{D}$ as: ${\tilde{\mathbf{x}}} = \mathcal{D}(\mathbf{z})$.

Stable diffusion formulates generation as a latent- space denoising task. A clean latent code \(\mathbf{z}\) is gradually corrupted by adding Gaussian noise via \(t\) steps, resulting in \(\mathbf{z}_t\). The model — a U-Net denoiser — is then trained to recover the original noise \(\epsilon\) from \(\mathbf{z}_t\) by minimising the following objective:

\begin{equation}
\label{eq1}
    \mathcal{L}_{\text{diffusion}} = \mathbb{E}_{{\mathbf{z}}_{t}, c, \epsilon, t} \left[ \left\| \epsilon - {\epsilon}_{\theta}({\mathbf{z}}_{t}, c, t) \right\|_2^2 \right],
\end{equation}

\noindent
where ${\epsilon}_{\theta}, c$ are the denoising UNet, and conditional driving the deoising process, respectively.

\begin{figure}[t!]
    \centering
    \begin{overpic}[width=.48\textwidth]{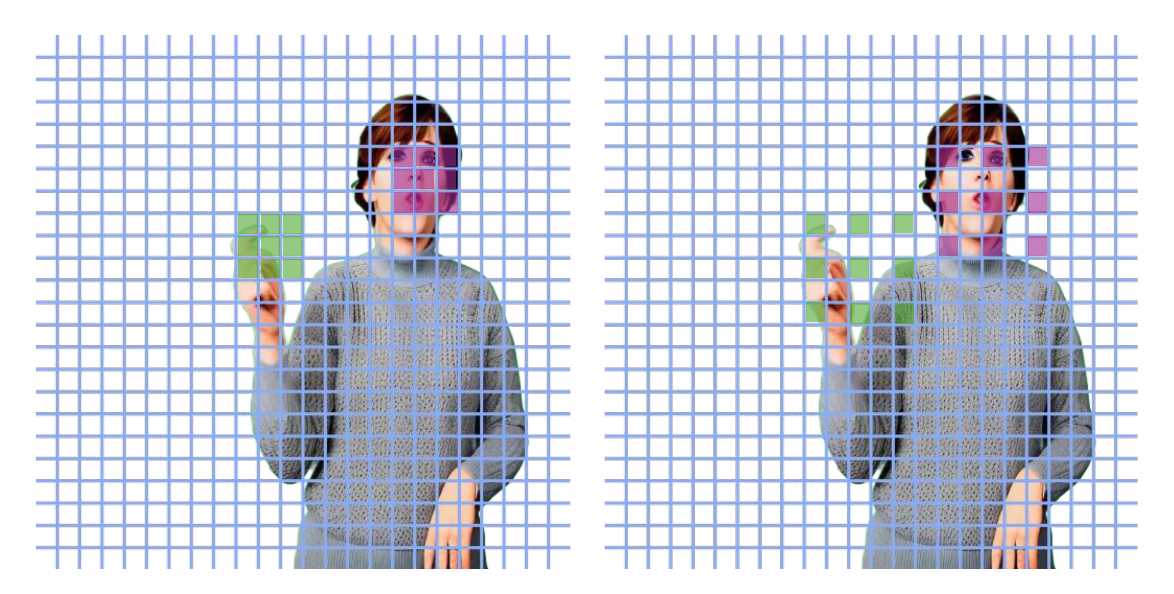}
    
    \put(25.0,-1){ $d = 1$}

    \put(73.0,-1){ $d = 2$}

  \end{overpic}
    \caption{\textbf{Sign features aggregation.} Visual representation motivating the use of convolution filters with dilations at various rate. This will allow our network to see the sign at various level of details to capture both manual and non-manual features.}
    \label{fig:psimotion}
\end{figure}

\subsection{Sign features aggregation}~\label{sec:aggregation}

We propose a novel aggregation module that effectively incorporates the manual and non-manual sign features into the denoising process. To this end, we propose a multi-scale spatio-temporal feature aggregation that preserves the content of the sign while being independent of the signer's identity. Such a representation makes it easy to model the sign and adapt it to different identities.

Given the input human pose, hand and face, we encode each of the inputs into a separate latent representation. Therefore, we get three features as: $\texttt{f}_{\text{pose}} = \mathcal{E}_{pose}(\mathbf{x}_{\text{pose}})$, $\texttt{f}_{\text{hand}} = \mathcal{E}_{hand}(\mathbf{x}_{\text{hand}})$, and $\texttt{f}_{\text{head}} = \mathcal{E}_{head}(\mathbf{x}_{\text{head}}) \in \mathbb{R}^{c \times t \times h \times w}$, where $c, t, h, w$ are the number of channels, time steps, width and height respectively. Note that there may be different implementations for each of the input encoders $\mathcal{E}_{\star}$, but we use a similar architecture to ControleNet~\cite{zhang2023adding}, which we find effective.

Our aggregation module $\Psi_{\text{motion}}$ first uses a multi-scale dilation convolution with filters at different dilation rates to capture both fine-grained (\textit{i.e.}, non-manual features) and notable actions (\textit{i.e.}, manual features) in the sign language. Dilation was used in high-resolution image generation to account for different spatial scales and contextual information so that the model can effectively integrate local details and global structures~\cite{du2024demofusion,he2023scalecrafter}. On this basis, we use dilation convolution at different scales to model manual and non-manual features. To do so, we apply 3D convolutional kernels with dilations at \(d \in \{1, 2, 4\}\) to each input feature.

For each tensor from \(\mathbf{f}_\text{pose}\), \(\mathbf{f}_\text{hand}\), or \(\mathbf{f}_\text{face}\), we apply 3D convolutional kernels with dilation rates \(d \in \{1, 2, 4\}\). The dilated convolution operation for a feature tensor \(\mathbf{h} \in \mathbb{R}^{t \times h \times w}\) (omitting channels for simplicity) with a kernel \(\mathbf{k} \in \mathbb{R}^{r \times r \times r}\) and dilation rate \(d\) is:

\begin{equation}
f_{\mathbf{k}}^d(\mathbf{h})(\mathbf{o}) = \sum_{\mathbf{t} \in \{-1, 0, 1\}^3} \mathbf{h}(\mathbf{o} + d \cdot \mathbf{t}) \cdot \mathbf{k}(\mathbf{t}),
\end{equation}

where \(\mathbf{o} = (o_t, o_h, o_w)\) denotes the output position, and \mbox{\(\mathbf{t} = (t_t, t_h, t_w)\)} represents the kernel offsets. For each dilation rate \(d\), we compute a multi-scale feature by summing the convolved outputs across all modalities:

\begin{equation}
\mathbf{m}_d = f_{\mathbf{k}}^d(\mathbf{f}_\text{pose}) + f_{\mathbf{k}}^d(\mathbf{f}_\text{hand}) + f_{\mathbf{k}}^d(\mathbf{f}_\text{face}).
\end{equation}

Next, we obtain the cross-feature information to model the interactions between manual and non-manual components from the pose, hand, and face modalities. The input features are concatenated along the channel dimension and processed with a $1 \times 1$ 3D convolution:
\begin{equation}
\mathbf{c} = \texttt{ReLU}\left( \texttt{conv}_{1\times1} \left( \texttt{concat}(\mathbf{f}_\text{pose}, \mathbf{f}_\text{hand}, \mathbf{f}_\text{face}) \right) \right).    
\end{equation}

The final feature is obtained by concatenating all intermediate representations and applying a fusion convolution, followed by a residual connection to enhance the training stability:
\begin{equation}
\mathbf{a} = \texttt{concat}\left( \mathbf{f}_\text{pose}, \mathbf{f}_\text{hand}, \mathbf{f}_\text{face}, \mathbf{c},  \mathbf{m}_1, \mathbf{m}_2, \mathbf{m}_3 \right),
\end{equation}
\begin{equation}
\Psi_{\text{motion}}(\mathbf{f}_\text{pose}, \mathbf{f}_\text{hand}, \mathbf{f}_\text{face}) = \texttt{conv}_{1\times1}(\mathbf{a}) + \frac{\mathbf{f}_\text{pose} + \mathbf{f}_\text{hand} + \mathbf{f}_\text{face}}{3}.
\end{equation}

\subsection{Architecture}\label{sub:architecture}
Our model architecture synthesises diverse, photorealistic sign language avatars using a diffusion-based framework that captures manual and non-manual features with high linguistic accuracy, while discarding identity-specific cues. To achieve this, we combine a ControlNet-based Conditional Latent Diffusion Backbone~\cite{zhang2023adding}, which provides structural control through pose and gesture conditioning, with a pre-trained visual foundation model that enriches the synthesis with semantic and photo-realistic details. This design enables the creation of visually diverse and inclusive avatars that reflect a wide range of human variations without being linked to real identities.

Our architecture leverages a latent diffusion backbone, using ControlNet~\cite{zhang2023adding} as an encoder for manual and non-manual features. As described in Equation~\ref{eq1}, the combined manual and non-manual features form the input condition \(c\) for the denoising network. Thanks to this construction, our aggregation module can be seamlessly adapted to any Latent Diffusion Model (LDM) architecture. In our implementation, we follow the structure proposed in~\cite{Wang_ACMM_24,10656410}, where a reference image is processed by a separate network that mirrors the architecture of the denoising network.

Starting from the input modalities —the pose maps \((\mathbf{x}_{\text{pose}})\), the hand masks \((\mathbf{x}_{\text{hand}})\) and the face priors \((\mathbf{x}_{\text{head}})\) — we extract modality-specific features using lightweight encoders: \(\texttt{f}_{\text{pose}} = \mathcal{E}_{\text{pose}}(\mathbf{x}_{\text{pose}})\), \(\texttt{f}_{\text{hand}} = \mathcal{E}_{\text{hand}}(\mathbf{x}_{\text{hand}})\), and \(\texttt{f}_{\text{head}} = \mathcal{E}_{\text{head}}(\mathbf{x}_{\text{head}})\). These representations are then fused by our custom aggregation module \(\Psi_{\text{motion}}\), which integrates manual and non-manual signals across spatial and temporal scales (Sec.~\ref{sec:aggregation}).

In parallel to our ControlNet-based encoders, we process the hand and face inputs (\(\mathbf{x}_{\text{hand}}\), \(\mathbf{x}_{\text{head}}\)) using a frozen visual foundation model. Foundation models~\cite{oquab2023dinov2,10.1007/978-3-031-73235-5_12,DBLP:journals/corr/abs-2103-00020} have proven their effectiveness in various visual tasks due to their ability to extract rich, generalisable features. In our case, we use Sapiens visual features~\cite{10.1007/978-3-031-73235-5_12}, \(\Psi_{\text{sapien}}\), which provide a powerful abstraction of the manual and non-manual features and allow the model to focus on the semantic content of the sign rather than identity-specific features.

The outputs of the sign aggregation module \(\Psi_{\text{motion}}\) and the foundation model \(\Psi_{\text{sapien}}\) are combined with encoded modality features to form the conditioning input $(c)$, which is then fed into the denoising UNet of the diffusion model. This conditioning input is defined as:

\begin{equation}
\begin{split}
c =\; & \mathbf{f}_\text{pose} + \mathbf{f}_\text{hand} + \mathbf{f}_\text{face} \;+\; \Psi_{\text{sapien}}\bigl(\mathbf{f}_\text{hand},\mathbf{f}_\text{face}\bigr) \\ 
& +\; \lambda\,\Psi_{\text{motion}}\bigl(\mathbf{f}_\text{pose},\mathbf{f}_\text{hand},\mathbf{f}_\text{face}\bigr)\,,
\end{split}
\end{equation}

where $\lambda$ is a controlling factor.
\section{Experiments}

\subsection{Dataset}
We train and evaluate our model on the YouTube-SL-25 dataset~\cite{tanzer2024youtubesl25largescaleopendomainmultilingual}, which contains a diverse collection of sign language videos from multiple signers. Following the preprocessing pipeline described in the supplementary, we detect and isolate the primary signer in each frame, apply matting to remove background clutter, and subsample up to 120 frames per video. This yields a total of 13,300 videos for training and 4,068 videos for testing. The data extraction was performed on a cluster of 40 NVIDIA RTX 3090 GPUs over a period of four days to process all images.

\subsection{Implementation Details}
Our model is based on the Stable Diffusion 2.1 backbone~\cite{9878449}, which has been extended to include pose, hand and face conditioning in addition to our aggregation and foundation feature modules. After~\cite{10656410} the training takes place in two phases. We train for 30,000 iterations with a batch size of 64 and use the AdamW optimiser (learning rate 1$e$-5, weight decay 1$e$-2) on four NVIDIA A100 GPUs. The diffusion process spans 1,000 time steps, with the early (high-noise) time steps capturing the global pose and motion structure and the later (low-noise) time steps refining the fine facial expressions and hand details. In the second phase, we train the temporal aggregation module~\cite{guo2023animatediff} for 10,000 iterations with 24-frame clips at a batch size of 2 so that the network can learn smooth motion transitions and maintain temporal consistency.

\subsection{Evaluation metrics}
To comprehensively evaluate our model, we use both objective metrics and human judgement.

\textbf{Quantitative Metrics.} We use three widely adopted measures to evaluate image quality. PSNR (Peak Signal-to-Noise Ratio) quantifies the pixel fidelity between the generated and the real images. Higher PSNR values indicate more accurate pixel reconstructions~\cite{Wang_2004_TIP}. SSIM (Structural Similarity Index), evaluates the preservation of structural and perceptual consistency, taking into account aspects such as luminance, contrast and texture~\cite{Wang_2004_TIP}. And finally, LPIPS (Learned Perceptual Image Patch Similarity) captures perceptual differences using deep features from pre-pretrained networks. Lower scores indicate higher perceptual similarity~\cite{zhang2018perceptual}. We deliberately omit video-level metrics such as Fréchet Video Distance (FVD)~\cite{unterthiner2018towards}, as they are primarily used to measure temporal coherence and visual realism between sequences. In our context, such metrics provide only limited insight into the accuracy or correctness of sign language articulation, which is crucial for this task.

\textbf{User Study.} In addition to the quantitative metrics, we conduct a user study in which participants are shown videos generated by different methods and asked to rate them based on visual realism and sign accuracy.

\subsection{State-of-the-art comparison}

\textbf{Baselines.} We compare our method against two strong state-of-the-art baselines. In particular, we compare it to PENet~\cite{10581951}, a VAE-based method that separates the appearance from the encoding of the pose. We also compare our model against MimicMotion~\cite{zhang2025mimicmotion}, ControlNext~\cite{peng2024controlnext}, UniAnimate-DiT~\cite{wang2025unianimate}, and AnimateAnyone~\cite{10656410}, which are categorised as LDM models.

\textbf{Quantitative comparison.} Table~\ref{tab:soa_comp} summarises the results of our method compared to PENet and AnimateAnyone. Our model consistently outperforms both baselines on all quantitative metrics. In particular, it achieves the highest PSNR value (22.71), clearly outperforming PENet (12.74), MimicMotion (17.79), ControlNext (17.87), UniAnimate-DiT (18.15), and AnimateAnyone (18.40), indicating a better quality of reconstruction at the pixel level. In terms of structural similarity, our method achieves a SSIM of 0.8635, outperforming PENet (0.7277), MimicMotion (0.8006), ControlNext (0.8042), UniAnimate-DiT (0.8132), and AnimateAnyone (0.8019). For perceptual similarity, our approach achieves the lowest LPIPS value (0.1139), reflecting a closer alignment with ground-truth images in the feature space than PENet (0.2439), MimicMotion (0.1765), ControlNext (0.1711), UniAnimate-DiT (0.1831), and AnimateAnyone (0.1677). These results emphasise the effectiveness of our model in producing high quality, structurally consistent and perceptually faithful results.

\begin{figure*}[t!]
    \centering
    \includegraphics[width=1.\linewidth]{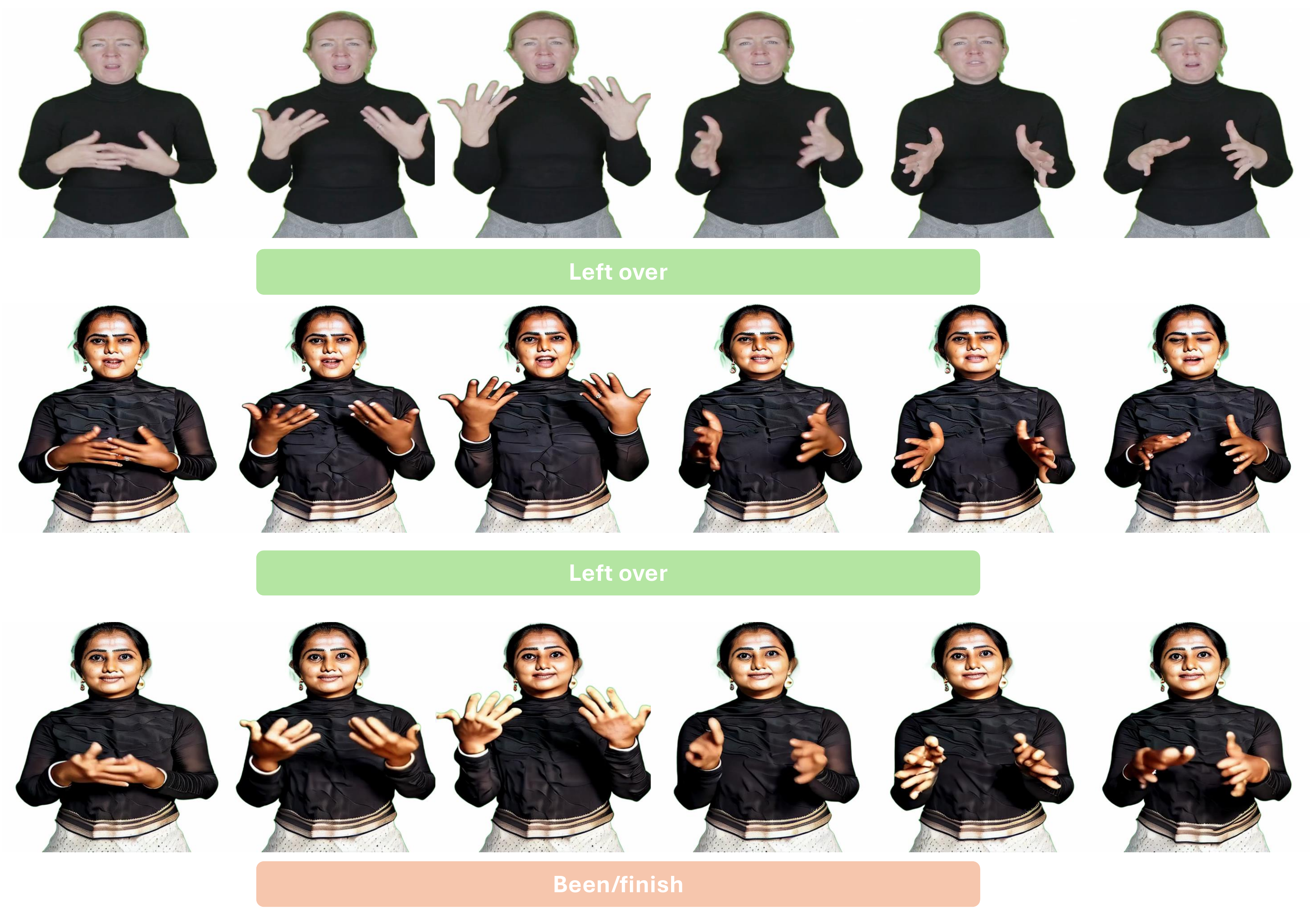}
    \caption{\textbf{Gloss-based synthesis.} Our model is able to generate glosses faithfully. In this example, we show that our proposed model is significantly better than AnimateAnyone~\cite{10656410}, where the non-manual features are not present and the hands do not convey the actual meaning of the sign. KEY -- (top row): Ground truth; (middle row): our method; (bottom row): AnimateAnyone.}
    \label{fig:gloss}
\end{figure*}

\begin{figure*}[t!]
    \centering
    \includegraphics[width=1.\linewidth]{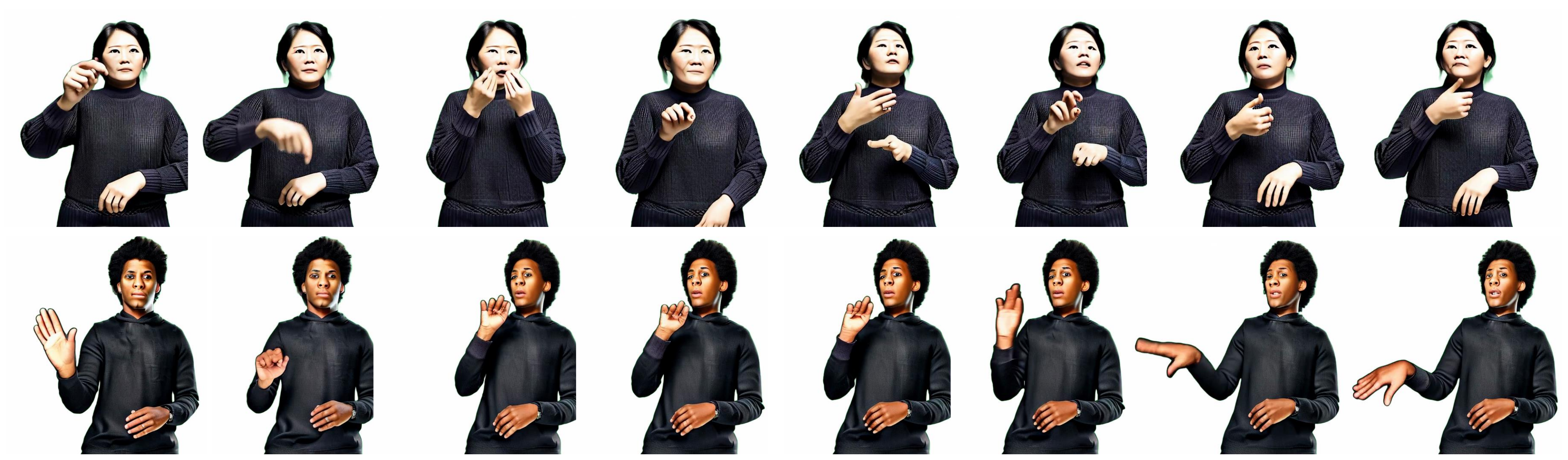}
    \caption{\textbf{Continuous sign.} Our model is able to generate continuous sign sequences. In this two examples, we show the results on Swiss-German Sign Language (DSGS) with two generated identities. }
    \label{fig:cnt}
\end{figure*}

\textbf{Qualitative Comparison.} Figure~\ref{fig:gloss} illustrates the misleading assumption that a visually coherent synthesis is acceptable. This is not the case in sign language, as it could completely change the meaning. As the user study shows, our model can faithfully model the non-manuals faithfully compared to previous methods. In addition, our model is able to generate long continuous sequences while maintaining visual fidelity and sign accuracy (Fig.~\ref{fig:cnt}).

\begin{table}[t!]
    \footnotesize
    \centering
    \caption{Comparison of different encoder configurations for appearance and feature extraction. }
    \label{tab:feat_app}
    \begin{tabular}{lccc}
    \hline
    \textbf{Method} & \textbf{PSNR} & \textbf{SSIM} & \textbf{LPIPS}  \\
    \hline
    $\mathcal{E}_{\texttt{app}} (\texttt{frozen})$ & 19.73 & 0.7914 & 0.1332 \\ 
    $\mathcal{E}_{\texttt{app}} (\texttt{Clip})$ & 20.87 & 0.8432 & 0.1327 \\ \midrule
    $\mathcal{E}_{\texttt{feat}} (\texttt{DINO}) $ & 20.82 & 0.8426 & 0.1328 \\
    $\mathcal{E}_{\texttt{feat}} (\texttt{ViT}) $ & \textbf{21.06} & \textbf{0.8458} & \textbf{0.1305} \\ \hline
    \end{tabular}
    
\end{table}


\begin{figure*}
    \centering
    \includegraphics[width=.8\linewidth]{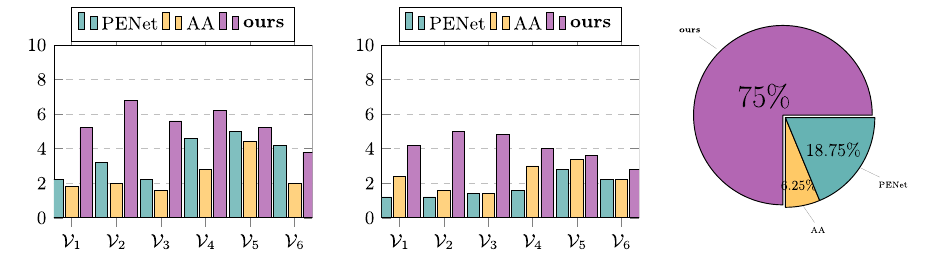}
    \caption{\textbf{User study.} We highlight the accuracy of the synthesised videos (left), the realism of the videos (centre) and user preference (right).}
    \label{fig:user_study}
\end{figure*}


\subsection{Ablation Study}
To isolate and quantify the effects of each component in our proposed pipeline, we perform the following ablation analysis. First, we compare different encoder backbones for both appearance and feature extraction. Then, we evaluate the standalone effect of our sign feature aggregation module (\(\psi_{\texttt{motion}}\)), followed by its integration with the pre-trained visual foundation model (\(\psi_{\texttt{sapien}}\)).

\textbf{Encoder Selection.}  We first analyse the effects of different encoders for appearance (\(\mathcal{E}_{\texttt{app}}\)) and feature extraction (\(\mathcal{E}_{\texttt{feat}}\)) (Tab.~\ref{tab:feat_app}). We first compare two variants of our appearance encoder, both based on CLIP: one with frozen weights $\mathcal{E}_{\texttt{app}} (\texttt{frozen})$. Freezing CLIP leads to PSNR = 19.73 dB and LPIPS = 0.1332, while fine-tuning increases PSNR to 20.87 dB and reduces LPIPS to 0.1327. This improvement emphasises the advantage of matching the rich representations of CLIP to our signer synthesis task.

For feature extraction, DINO (\(\mathcal{E}_{\texttt{feat}} (\texttt{DINO})\)) achieves PSNR=20.82dB, SSIM=0.8426, LPIPS=0.1328, while a Vision Transformer \(\mathcal{E}_{\texttt{feat}} (\texttt{ViT})\) further improves the performance to PSNR=21.06dB, SSIM=0.8458, LPIPS=0.1305. We therefore adopt CLIP for appearance encoding and ViT for feature extraction as our baseline.

\textbf{Impact of Motion and Sapien Modules.} Next, we evaluate the contributions of the sign feature aggregation module (\(\psi_{\texttt{motion}}\)) and the Sapien module (\(\psi_{\texttt{sapien}}\)) (Tab.~\ref{ref:motion_sapiens}). Starting from the CLIP+ViT baseline (PSNR=21.06, SSIM=0.8458, LPIPS=0.1305), we gradually add our aggregation and foundation modules. Including only the aggregation module \(\psi_{\texttt{motion}}\) (Sec.~\ref{sec:aggregation}) leads to a PSNR gain of 21.36 dB, but SSIM drops to 0.8397 and LPIPS increases to 0.1329, suggesting that the aggregation module without Sapiens features may overemphasise structural consistency at the expense of visual fidelity. In contrast, adding \(\psi_{\texttt{motion}}\) and the pre-trained Sapien module \(\psi_{\texttt{sapien}}\) (Sec.~\ref{sub:architecture}) dramatically boosts performance to PSNR=22.71dB, SSIM=0.8635 and LPIPS=0.1139. The Sapien features provide a rich semantic context that complements motion dynamics at multiple scales, leading to significant improvements in reconstruction and perceptual quality.

\textbf{Effect of \(\lambda\) in Motion Module.} We investigate how the weighting factor \(\lambda\) influences the contribution of \(\psi_{\texttt{motion}}\) (Tab.~\ref{ref:lambda}).  As \(\lambda\) increases, we observe a consistent increase in PSNR — from 21.33 dB at \(\lambda=1.0\) to 21.36 dB at \(\lambda=0.01\) — with only minimal variation in LPIPS and SSIM. We therefore retain (\(\lambda=0.01\)) as the default value.

\subsection{User Study}

To evaluate the effectiveness of our proposed method in generating realistic and semantically correct sign language videos, we conducted a user study comparing our approach with PENet~\cite{10581951} and AnimateAnyone~\cite{10656410}.~\footnote{Due to the time and effort required to evaluate all models, we selected the best-performing model among the LDM-based approaches, AnimateAnyone~\cite{10656410}.} We selected six different British Sign Language (BSL) sequences, that we noted as: $\mathcal{V}_1, \dots, \mathcal{V}_6$. Each sequence was synthesised using all three methods, for a total of 18 videos per participant.

We asked five native BSL interpreters to rate each video on two levels: \textbf{signing accuracy} (how correctly the sign conveys the intended message) and \textbf{visual realism} (how natural and lifelike the movement and synthesis appear). Ratings were recorded on a 10-point scale and videos were presented in random order to minimise bias.

Across all videos and participants, our method achieved the highest average scores in both metrics: sign accuracy of 5.47 ($\pm 2.11$) and visual realism of 4.07 ($\pm 2.21$). For comparison: AnimateAnyone scored 2.43 ($\pm 1.61$) for accuracy and 2.33 ($\pm 1.86$) for realism, while PENet scored 3.57 ($\pm 2.24$) for accuracy and 1.73 ($\pm 1.11$) for realism.

Our method was consistently better than the baselines in all six sign sequences. For example, for the sequence $\mathcal{V}_2$, our method scored 6.8 ($\pm$1.79) for accuracy and 5.0 ($\pm$2.74) for realism, while AnimateAnyone and PENet scored much lower. Even for challenging examples such as $\mathcal{V}_5$ and $\mathcal{V}_6$, our approach remained superior for perceptual accuracy.

Perhaps the most interesting result is that the aesthetic appeal for non-signers does not align with sign accuracy. For example, AnimateAnyone scores higher on the general visual quality metrics, but all native signers in BSL unanimously rated PENet as more accurate in conveying the intended signs. This discrepancy demonstrates the importance of involving members of the Deaf community throughout the model development and evaluation process. Their feedback ensures that the synthesised videos are not only visually appealing, but also convey the correct meaning of the signs.

\begin{table}[t!]
    \footnotesize
    \centering
    \caption{Ablation study showing the effect of motion and sapien modules.}
    \label{ref:motion_sapiens}
    \begin{tabular}{lccc}
    \hline
    \textbf{Method} & \textbf{PSNR} & \textbf{SSIM} & \textbf{LPIPS}  \\
    \hline
    baseline & 21.06 & 0.8458 & 0.1305 \\ \midrule
    $ + \psi_{\texttt{motion}} $ & 21.36 & 0.8397 & 0.1329 \\
    $ + \psi_{\texttt{motion}} + \psi_{\texttt{sapien}} $ & \textbf{22.71} & \textbf{0.8635} & \textbf{0.1139} \\ \hline
    \end{tabular}
    
\end{table}

\begin{table}[t!]
    \footnotesize
    \centering
    \caption{Effect of varying the weighting factor \(\lambda\) for the sign features aggregation module \(\psi_{\texttt{motion}}\).}
    \label{ref:lambda}
    \begin{tabular}{lccc}
    \hline
    \textbf{Method} & \textbf{PSNR} & \textbf{SSIM} & \textbf{LPIPS}  \\
    \hline
    w/o $ \psi_{\texttt{motion}} $  & 21.06 & \textbf{0.8458} & \textbf{0.1305} \\ \midrule
    $ + \psi_{\texttt{motion}} (\lambda = 1.0)$ & 21.33 & 0.8386 & 0.1336 \\
    $ + \psi_{\texttt{motion}} (\lambda = 0.1)$ & 21.34 & 0.8388 & 0.1333 \\
    $ + \psi_{\texttt{motion}} (\lambda = 0.01)$ & \textbf{21.36} & 0.8397 & 0.1329   \\ \hline
    \end{tabular}
    
\end{table}

\begin{table}[t!]
    \footnotesize
    \centering
        \caption{Quantitative comparison with state-of-the-art methods. We use reconstruction (PSNR, SSIM)~\cite{Wang_2004_TIP} and perceptual (LPIPS)~\cite{zhang2018perceptual} metrics. The best results are shown in \textbf{bold}.}
        \label{tab:soa_comp}
    \begin{tabular}{lccc}
    \hline
    \textbf{Method} & \textbf{PSNR} & \textbf{SSIM} & \textbf{LPIPS}  \\
    \hline
    PENet~\cite{10581951} & 12.74 & 0.7277 & 0.2439 \\
    MimicMotion~\cite{zhang2025mimicmotion} & 17.79 & 0.8006 & 0.1765 \\
    ControlNext~\cite{peng2024controlnext} & 17.87 & 0.8042 & 0.1711 \\
    UniAnimate-DiT~\cite{wang2025unianimate} & 18.15 & 0.8132 & 0.1831 \\
    AnimateAnyone~\cite{10656410}  & 18.40 & 0.8019 & 0.1677 \\ \hdashline
    Ours & \textbf{22.71} & \textbf{0.8635} & \textbf{0.1139} \\ \hline
    \end{tabular}
    
\end{table}

\section{Conclusion}
In this paper, we present a novel framework for photorealistic digital signer avatars that synthesises signers from diverse ethnic backgrounds. We propose to explicitly model the manual and non-manual features present in sign language through a novel feature aggregation module that applies dilation kernels at different rates to capture both fine-grained and coarse details in the input modalities. In addition, we leverage visual foundation model that provides rich semantic features to contribute to higher visual fidelity. Comprehensive quantitative evaluations against state-of-the-art baseline solutions show consistent gain from previous solutions.

Our model is able to synthesise both glosses and continuous sign sequences. The results of our user study suggest that feedback loops from the deaf community are essential, as relying on traditional metrics is not sufficient.

{
    \small
    \bibliographystyle{ieeenat_fullname}
    \bibliography{main}

\begin{thebibliography}{49}
\providecommand{\natexlab}[1]{#1}
\providecommand{\url}[1]{\texttt{#1}}
\expandafter\ifx\csname urlstyle\endcsname\relax
  \providecommand{\doi}[1]{doi: #1}\else
  \providecommand{\doi}{doi: \begingroup \urlstyle{rm}\Url}\fi

\bibitem[Alp~Güler et~al.(2018)Alp~Güler, Neverova, and Kokkinos]{Guler_2018_CVPR}
Rıza Alp~Güler, Natalia Neverova, and Iasonas Kokkinos.
\newblock {DensePose: Dense Human Pose Estimation in the Wild}.
\newblock In \emph{CVPR}, pages 7297--7306, 2018.

\bibitem[Arjovsky et~al.(2017)Arjovsky, Chintala, and Bottou]{pmlr-v70-arjovsky17a}
Martin Arjovsky, Soumith Chintala, and L{\'e}on Bottou.
\newblock {W}asserstein generative adversarial networks.
\newblock In \emph{Proceedings of the 34th International Conference on Machine Learning}, pages 214--223. PMLR, 2017.

\bibitem[Bhunia et~al.(2023)Bhunia, Khan, Cholakkal, Anwer, Laaksonen, Shah, and Khan]{bhunia2022pidm}
Ankan~Kumar Bhunia, Salman Khan, Hisham Cholakkal, Rao~Muhammad Anwer, Jorma Laaksonen, Mubarak Shah, and Fahad~Shahbaz Khan.
\newblock Person image synthesis via denoising diffusion model.
\newblock \emph{CVPR}, 2023.

\bibitem[Blei et~al.(2017)Blei, Kucukelbir, and McAuliffe]{Blei_2017}
David~M. Blei, Alp Kucukelbir, and Jon~D. McAuliffe.
\newblock Variational inference: A review for statisticians.
\newblock In \emph{Journal of the American Statistical Association}, 2017.

\bibitem[Dong et~al.(2018)Dong, Liang, Gong, Lai, Zhu, and Yin]{Dong_2018_NIPS}
Haoye Dong, Xiaodan Liang, Ke Gong, Hanjiang Lai, Jia Zhu, and Jian Yin.
\newblock {Soft-Gated Warping-GAN for Pose-Guided Person Image Synthesis}.
\newblock In \emph{NeurIPS}, 2018.

\bibitem[Du et~al.(2024)Du, Chang, Hospedales, Song, and Ma]{du2024demofusion}
Ruoyi Du, Dongliang Chang, Timothy Hospedales, Yi-Zhe Song, and Zhanyu Ma.
\newblock Demofusion: Democratising high-resolution image generation with no \$\$\$.
\newblock In \emph{CVPR}, 2024.

\bibitem[Goodfellow et~al.(2014)Goodfellow, Pouget-Abadie, Mirza, Xu, Warde-Farley, Ozair, Courville, and Bengio]{goodfellow2014generative}
Ian~J. Goodfellow, Jean Pouget-Abadie, Mehdi Mirza, Bing Xu, David Warde-Farley, Sherjil Ozair, Aaron Courville, and Yoshua Bengio.
\newblock {Generative Adversarial Networks}.
\newblock In \emph{NeurIPS}, 2014.

\bibitem[Guo et~al.(2024)Guo, Yang, Rao, Liang, Wang, Qiao, Agrawala, Lin, and Dai]{guo2023animatediff}
Yuwei Guo, Ceyuan Yang, Anyi Rao, Zhengyang Liang, Yaohui Wang, Yu Qiao, Maneesh Agrawala, Dahua Lin, and Bo Dai.
\newblock Animatediff: Animate your personalized text-to-image diffusion models without specific tuning.
\newblock \emph{International Conference on Learning Representations}, 2024.

\bibitem[Han et~al.(2023)Han, Zhu, Deng, Song, and Xiang]{10377856}
Xiao Han, Xiatian Zhu, Jiankang Deng, Yi-Zhe Song, and Tao Xiang.
\newblock Controllable person image synthesis with pose-constrained latent diffusion.
\newblock In \emph{2023 IEEE/CVF International Conference on Computer Vision (ICCV)}, pages 22711--22720, 2023.

\bibitem[He et~al.(2024)He, Yang, Chen, Cun, Xia, Zhang, Wang, He, Chen, and Shan]{he2023scalecrafter}
Yingqing He, Shaoshu Yang, Haoxin Chen, Xiaodong Cun, Menghan Xia, Yong Zhang, Xintao Wang, Ran He, Qifeng Chen, and Ying Shan.
\newblock Scalecrafter: Tuning-free higher-resolution visual generation with diffusion models.
\newblock In \emph{ICLR}, 2024.

\bibitem[Ho et~al.(2020)Ho, Jain, and Abbeel]{Ho_neurips_20}
Jonathan Ho, Ajay Jain, and Pieter Abbeel.
\newblock Denoising diffusion probabilistic models.
\newblock In \emph{Proceedings of the 34th International Conference on Neural Information Processing Systems}, Red Hook, NY, USA, 2020. Curran Associates Inc.

\bibitem[Hu et~al.(2024{\natexlab{a}})Hu, Fan, Wu, Xi, Lee, Pavlakos, and Wang]{hu2024expressive}
Hezhen Hu, Zhiwen Fan, Tianhao Wu, Yihan Xi, Seoyoung Lee, Georgios Pavlakos, and Zhangyang Wang.
\newblock Expressive gaussian human avatars from monocular {RGB} video.
\newblock In \emph{NeurIPS}, 2024{\natexlab{a}}.

\bibitem[Hu et~al.(2024{\natexlab{b}})Hu, Gao, Zhang, Sun, Zhang, and Bo]{10656410}
Li Hu, Xin Gao, Peng Zhang, Ke Sun, Bang Zhang, and Liefeng Bo.
\newblock Animate anyone: Consistent and controllable image-to-video synthesis for character animation.
\newblock In \emph{2024 IEEE/CVF Conference on Computer Vision and Pattern Recognition (CVPR)}, pages 8153--8163, 2024{\natexlab{b}}.

\bibitem[Hwang et~al.(2021)Hwang, Kim, and Park]{hwang2021non}
Eui~Jun Hwang, Jung-Ho Kim, and Jong~C. Park.
\newblock Non-autoregressive sign language production with gaussian space.
\newblock In \emph{The 32nd British Machine Vision Conference (BMVC 21)}. British Machine Vision Conference (BMVC), 2021.

\bibitem[Karpouzis et~al.(2007)Karpouzis, Caridakis, Fotinea, and Efthimiou]{KARPOUZIS200754}
K. Karpouzis, G. Caridakis, S.-E. Fotinea, and E. Efthimiou.
\newblock Educational resources and implementation of a greek sign language synthesis architecture.
\newblock \emph{Computers $\&$ Education}, 49\penalty0 (1):\penalty0 54--74, 2007.
\newblock Web3D Technologies in Learning, Education and Training.

\bibitem[Khirodkar et~al.(2024)Khirodkar, Bagautdinov, Martinez, Zhaoen, James, Selednik, Anderson, and Saito]{10.1007/978-3-031-73235-5_12}
Rawal Khirodkar, Timur Bagautdinov, Julieta Martinez, Su Zhaoen, Austin James, Peter Selednik, Stuart Anderson, and Shunsuke Saito.
\newblock Sapiens: Foundation for human vision models.
\newblock In \emph{ECCV}, pages 206--228, Cham, 2024. Springer Nature Switzerland.

\bibitem[Kim et~al.(2022)Kim, Kim, Lee, Cha, Lee, and Kim]{kim2022revisiting}
Taehun Kim, Kunhee Kim, Joonyeong Lee, Dongmin Cha, Jiho Lee, and Daijin Kim.
\newblock Revisiting image pyramid structure for high resolution salient object detection.
\newblock In \emph{ACCV}, 2022.

\bibitem[Lakhal and Bowden(2024)]{10581951}
Mohamed~Ilyes Lakhal and Richard Bowden.
\newblock Diversity-aware sign language production through a pose encoding variational autoencoder.
\newblock In \emph{2024 IEEE 18th International Conference on Automatic Face and Gesture Recognition (FG)}, pages 1--10, 2024.

\bibitem[Lakhal and Bowden(2025)]{Lakhal_FG_25}
Mohamed~Ilyes Lakhal and Richard Bowden.
\newblock Gaussiangan: Real-time photorealistic controllable human avatars.
\newblock In \emph{2025 IEEE 19th International Conference on Automatic Face and Gesture Recognition (FG)}, 2025.

\bibitem[Li et~al.(2019)Li, Huang, and Loy]{Li_2019_CVPR}
Yining Li, Chen Huang, and Chen~Change Loy.
\newblock {Dense Intrinsic Appearance Flow for Human Pose Transfer}.
\newblock In \emph{CVPR}, pages 3688--3697, 2019.

\bibitem[Loper et~al.(2015)Loper, Mahmood, Romero, Pons-Moll, and Black]{Loper_2015_TOG}
Matthew Loper, Naureen Mahmood, Javier Romero, Gerard Pons-Moll, and Michael~J. Black.
\newblock {SMPL: A Skinned Multi-Person Linear Model}.
\newblock \emph{ACM TOG}, 34\penalty0 (6):\penalty0 248:1--248:16, 2015.

\bibitem[Lu et~al.(2024)Lu, Zhang, Ma, Xie, and Lai]{10656324}
Yanzuo Lu, Manlin Zhang, Andy~J Ma, Xiaohua Xie, and Jianhuang Lai.
\newblock Coarse-to-fine latent diffusion for pose-guided person image synthesis.
\newblock In \emph{2024 IEEE/CVF Conference on Computer Vision and Pattern Recognition (CVPR)}, pages 6420--6429, 2024.

\bibitem[Ma et~al.(2017)Ma, Jia, Sun, Schiele, Tuytelaars, and Van~Gool]{Ma_2017_NIPS}
Liqian Ma, Xu Jia, Qianru Sun, Bernt Schiele, Tinne Tuytelaars, and Luc Van~Gool.
\newblock {Pose Guided Person Image Generation}.
\newblock In \emph{NeurIPS}, 2017.

\bibitem[Neverova et~al.(2018)Neverova, Alp~Guler, and Kokkinos]{Neverova_2018_ECCV}
Natalia Neverova, Riza Alp~Guler, and Iasonas Kokkinos.
\newblock {Dense Pose Transfer}.
\newblock In \emph{ECCV}, pages 128--143, 2018.

\bibitem[Oquab et~al.(2023)Oquab, Darcet, Moutakanni, Vo, Szafraniec, Khalidov, Fernandez, Haziza, Massa, El-Nouby, Howes, Huang, Xu, Sharma, Li, Galuba, Rabbat, Assran, Ballas, Synnaeve, Misra, Jegou, Mairal, Labatut, Joulin, and Bojanowski]{oquab2023dinov2}
Maxime Oquab, Timothée Darcet, Theo Moutakanni, Huy~V. Vo, Marc Szafraniec, Vasil Khalidov, Pierre Fernandez, Daniel Haziza, Francisco Massa, Alaaeldin El-Nouby, Russell Howes, Po-Yao Huang, Hu Xu, Vasu Sharma, Shang-Wen Li, Wojciech Galuba, Mike Rabbat, Mido Assran, Nicolas Ballas, Gabriel Synnaeve, Ishan Misra, Herve Jegou, Julien Mairal, Patrick Labatut, Armand Joulin, and Piotr Bojanowski.
\newblock Dinov2: Learning robust visual features without supervision, 2023.

\bibitem[Peng et~al.(2024)Peng, Wang, Zhang, Li, Yang, and Jia]{peng2024controlnext}
Bohao Peng, Jian Wang, Yuechen Zhang, Wenbo Li, Ming-Chang Yang, and Jiaya Jia.
\newblock Controlnext: Powerful and efficient control for image and video generation.
\newblock \emph{arXiv preprint arXiv:2408.06070}, 2024.

\bibitem[Radford et~al.(2021)Radford, Kim, Hallacy, Ramesh, Goh, Agarwal, Sastry, Askell, Mishkin, Clark, Krueger, and Sutskever]{DBLP:journals/corr/abs-2103-00020}
Alec Radford, Jong~Wook Kim, Chris Hallacy, Aditya Ramesh, Gabriel Goh, Sandhini Agarwal, Girish Sastry, Amanda Askell, Pamela Mishkin, Jack Clark, Gretchen Krueger, and Ilya Sutskever.
\newblock Learning transferable visual models from natural language supervision.
\newblock \emph{CoRR}, abs/2103.00020, 2021.

\bibitem[Rombach et~al.(2022)Rombach, Blattmann, Lorenz, Esser, and Ommer]{9878449}
Robin Rombach, Andreas Blattmann, Dominik Lorenz, Patrick Esser, and Björn Ommer.
\newblock High-resolution image synthesis with latent diffusion models.
\newblock In \emph{2022 IEEE/CVF Conference on Computer Vision and Pattern Recognition (CVPR)}, pages 10674--10685, 2022.

\bibitem[Sandler and Lillo-Martin(2006)]{Sandler_Lillo-Martin_2006}
Wendy Sandler and Diane Lillo-Martin.
\newblock \emph{Sign Language and Linguistic Universals}.
\newblock Cambridge University Press, 2006.

\bibitem[Saunders et~al.(2020)Saunders, Camgoz, and Bowden]{saunders2020progressive}
Ben Saunders, Necati~Cihan Camgoz, and Richard Bowden.
\newblock {Progressive Transformers for End-to-End Sign Language Production}.
\newblock In \emph{Proceedings of the European Conference on Computer Vision (ECCV)}, 2020.

\bibitem[Saunders et~al.(2021{\natexlab{a}})Saunders, Camgoz, and Bowden]{Saunders_2021_FG}
Ben Saunders, Necati~Cihan Camgoz, and Richard Bowden.
\newblock {Anonysign: Novel Human Appearance Synthesis for Sign Language Video Anonymisation}.
\newblock In \emph{2021 16th IEEE International Conference on Automatic Face and Gesture Recognition (FG 2021)}, 2021{\natexlab{a}}.

\bibitem[Saunders et~al.(2021{\natexlab{b}})Saunders, Camgoz, and Bowden]{saunders2021continuous}
Ben Saunders, Necati~Cihan Camgoz, and Richard Bowden.
\newblock {Continuous 3D Multi-Channel Sign Language Production via Progressive Transformers and Mixture Density Networks}.
\newblock In \emph{International Journal of Computer Vision (IJCV)}, 2021{\natexlab{b}}.

\bibitem[Shen and Tang(2024)]{shen2024imagpose}
Fei Shen and Jinhui Tang.
\newblock Imagpose: A unified conditional framework for pose-guided person generation.
\newblock In \emph{The Thirty-eighth Annual Conference on Neural Information Processing Systems}, 2024.

\bibitem[Shen et~al.(2024)Shen, Ye, Zhang, Wang, Han, and Yang]{shen2024advancingposeguidedimagesynthesis}
Fei Shen, Hu Ye, Jun Zhang, Cong Wang, Xiao Han, and Wei Yang.
\newblock Advancing pose-guided image synthesis with progressive conditional diffusion models.
\newblock In \emph{ICLR}, 2024.

\bibitem[Silveira et~al.(2022)Silveira, Alaniz, Hurtado, Da~Silva, and De~Bem]{Silveira_2022_SIBGRAPI}
Wellington Silveira, Andrew Alaniz, Marina Hurtado, Bernardo~Castello Da~Silva, and Rodrigo De~Bem.
\newblock {SynLibras: A Disentangled Deep Generative Model for Brazilian Sign Language Synthesis}.
\newblock In \emph{2022 35th SIBGRAPI Conference on Graphics, Patterns and Images (SIBGRAPI)}, pages 210--215, 2022.

\bibitem[Tamura and Kawasaki(1988)]{Tamura_88}
S. Tamura and S. Kawasaki.
\newblock Recognition of sign language motion images.
\newblock \emph{Pattern Recogn.}, 21\penalty0 (4):\penalty0 343–353, 1988.

\bibitem[Tanzer and Zhang(2024)]{tanzer2024youtubesl25largescaleopendomainmultilingual}
Garrett Tanzer and Biao Zhang.
\newblock Youtube-sl-25: A large-scale, open-domain multilingual sign language parallel corpus, 2024.

\bibitem[{ultralytics}(2024)]{yolo11s}
{ultralytics}.
\newblock ultralytics.
\newblock \url{https://github.com/ultralytics/ultralytics}, 2024.
\newblock Accessed: 2025-08-20.

\bibitem[Unterthiner et~al.(2018)Unterthiner, van Steenkiste, Kurach, Marinier, Michalski, and Gelly]{unterthiner2018towards}
Thomas Unterthiner, Sjoerd van Steenkiste, Karol Kurach, Raphael Marinier, Marcin Michalski, and Sylvain Gelly.
\newblock Towards accurate generative models of video: A new metric \& challenges.
\newblock \emph{arXiv preprint arXiv:1812.01717}, 2018.

\bibitem[Wang et~al.(2025)Wang, Zhang, Gao, Wang, Zhou, Zhang, Yan, and Sang]{wang2025unianimate}
Xiang Wang, Shiwei Zhang, Changxin Gao, Jiayu Wang, Xiaoqiang Zhou, Yingya Zhang, Luxin Yan, and Nong Sang.
\newblock Unianimate: Taming unified video diffusion models for consistent human image animation.
\newblock \emph{Science China Information Sciences}, 2025.

\bibitem[Wang et~al.(2024)Wang, Dai, Chan, Zhou, Zhang, and Liu]{Wang_ACMM_24}
Yuanbin Wang, Weilun Dai, Long Chan, Huanyu Zhou, Aixi Zhang, and Si Liu.
\newblock Gpd-vvto: Preserving garment details in video virtual try-on.
\newblock In \emph{Proceedings of the 32nd ACM International Conference on Multimedia}, page 7133–7142, New York, NY, USA, 2024. Association for Computing Machinery.

\bibitem[Wang et~al.(2004)Wang, Bovik, Sheikh, and Simoncelli]{Wang_2004_TIP}
Zhou Wang, A.~C. Bovik, H.~R. Sheikh, and E.~P. Simoncelli.
\newblock {Image Quality Assessment: From Error Visibility to Structural Similarity}.
\newblock \emph{IEEE TIP}, 13\penalty0 (4):\penalty0 600--612, 2004.

\bibitem[Wold et~al.(1987)Wold, Esbensen, and Geladi]{WOLD198737}
Svante Wold, Kim Esbensen, and Paul Geladi.
\newblock Principal component analysis.
\newblock \emph{Chemometrics and Intelligent Laboratory Systems}, 2\penalty0 (1):\penalty0 37--52, 1987.

\bibitem[Yang and Lee(2011)]{6016973}
Hee-Deok Yang and Seong-Whan Lee.
\newblock Combination of manual and non-manual features for sign language recognition based on conditional random field and active appearance model.
\newblock In \emph{2011 International Conference on Machine Learning and Cybernetics}, pages 1726--1731, 2011.

\bibitem[Zelinka and Kanis(2020)]{9093516}
Jan Zelinka and Jakub Kanis.
\newblock Neural sign language synthesis: Words are our glosses.
\newblock In \emph{2020 IEEE Winter Conference on Applications of Computer Vision (WACV)}, pages 3384--3392, 2020.

\bibitem[Zhang et~al.(2025{\natexlab{a}})Zhang, Liu, Lin, Zhu, Li, Qin, Li, and Wang]{zhang2025guava}
Dongbin Zhang, Yunfei Liu, Lijian Lin, Ye Zhu, Yang Li, Minghan Qin, Yu Li, and Haoqian Wang.
\newblock {GUAVA: Generalizable Upper Body 3D Gaussian Avatar}.
\newblock In \emph{IEEE/CVF International Conference on Computer Vision (ICCV)}, 2025{\natexlab{a}}.

\bibitem[Zhang et~al.(2023)Zhang, Rao, and Agrawala]{zhang2023adding}
Lvmin Zhang, Anyi Rao, and Maneesh Agrawala.
\newblock Adding conditional control to text-to-image diffusion models, 2023.

\bibitem[Zhang et~al.(2018)Zhang, Isola, Efros, Shechtman, and Wang]{zhang2018perceptual}
Richard Zhang, Phillip Isola, Alexei~A Efros, Eli Shechtman, and Oliver Wang.
\newblock {The Unreasonable Effectiveness of Deep Features as a Perceptual Metric}.
\newblock In \emph{CVPR}, 2018.

\bibitem[Zhang et~al.(2025{\natexlab{b}})Zhang, Gu, Wang, Wang, Cheng, Zhu, and Zou]{zhang2025mimicmotion}
Yuang Zhang, Jiaxi Gu, Li-Wen Wang, Han Wang, Junqi Cheng, Yuefeng Zhu, and Fangyuan Zou.
\newblock Mimicmotion: High-quality human motion video generation with confidence-aware pose guidance.
\newblock In \emph{International Conference on Machine Learning}, 2025{\natexlab{b}}.

\end{thebibliography}
}
\clearpage
\setcounter{page}{1}
\maketitlesupplementary

\section{Sign Dataset}~\label{sub:train_dataset}

To effectively train our model, we propose a heuristic pipeline to prepare the data before extracting the modalities. We found that this step is an essential part to allow our model to focus on the non-manual features, which are bottlenecks to effectively synthesise them. The annotation process is detailed below.

\subsection{Preprocess} Our preprocessing pipeline is designed to isolate and extract the key modalities that are important for capturing manual (\textit{e.g.}, hand gestures) and non-manual (\textit{e.g.}, facial expressions) features. To avoid redundancy, we take a random subsample
$N = 120$ different images per video. The pipeline starts with person detection using the YOLOv11s~\cite{yolo11s} model, which identifies and retains only the largest person in each frame using the bounding box area, while masking out other detected persons to focus on the main signer. Following this, a matting model separates the signer from the background using the method proposed in~\cite{kim2022revisiting}.

\subsection{Modalities} To improve the representation of the manual features, we apply principal component analysis (PCA)~\cite{WOLD198737} to remove the dependencies of skin colour during the denoising process. More specifically, for a hand image with dimensions \( h \times w \times c \), we first reshape the image into a matrix of the form \( (h \cdot w) \times c \), treating each pixel as a \( c \)-dimensional feature vector. PCA is then applied to project these vectors onto a single principal component, effectively reducing the feature dimension to one per pixel. We have empirically found that this compact representation helps to model manual features efficiently.

We use bounding boxes of facial features to preserve only the eyes and mouth. We believe that this preserves the important information to correctly model the non-manual features while removing the identity information. During training, the model learns to use the bounding box to drive the reference image.

\begin{figure}[t!]
    \centering
    \includegraphics[width=1.\linewidth]{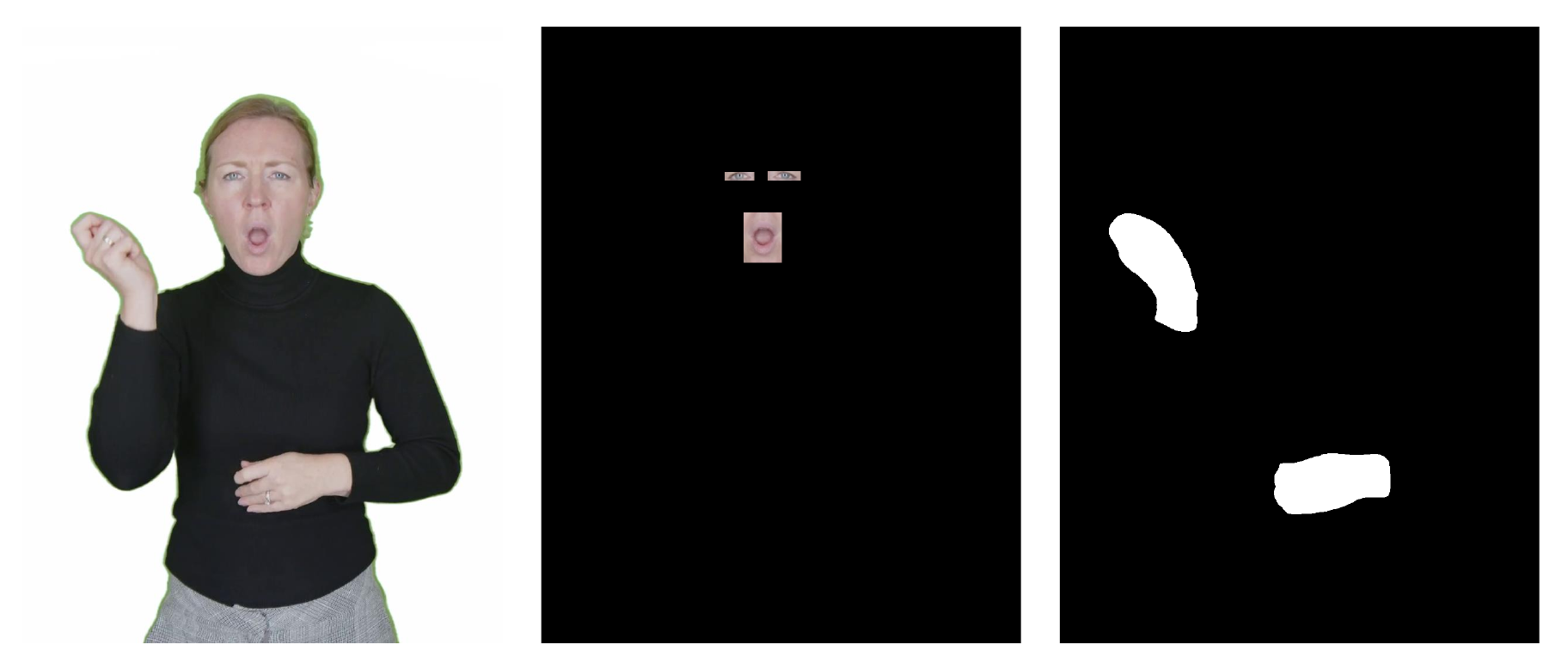}
    \caption{\textbf{Dataset modalities.} For each image, we extract non-manual (centre) and manual (right) cues to explicitly model them in our framework.}
    \label{fig:mods}
\end{figure}

\end{document}